%% file: main.tex
\documentclass{article}

\usepackage{spconf, amsmath, graphicx}
\usepackage{times}
\usepackage{epsfig}
\usepackage{amssymb}
\usepackage{qiangstyle}
\usepackage{multirow}
\usepackage[nobiblatex]{xurl}

\renewcommand{\paragraph}[1]{\noindent\textbf{#1}\quad}

\title{Noisy training improves E2E ASR for the edge}
\tname{
Dilin Wang, Yuan Shangguan, Haichuan Yang, Pierce Chuang}{
Jiatong Zhou, Meng Li, Ganesh Venkatesh, Ozlem Kalinli, Vikas Chandra
}

\address{Facebook AI}

\begin{document}

\maketitle

\input{tex/abstract}

\input{tex/intro}

\input{tex/method}

\input{tex/exp}

\input{tex/emformer}

\input{tex/priorwork}

\input{tex/conclusion}

\bibliographystyle{IEEEbib}
\bibliography{asr}

\end{document}

%% file: tex/abstract.tex
\begin{abstract}
Automatic speech recognition (ASR) has become increasingly ubiquitous on modern edge devices. Past work developed streaming End-to-End (E2E) all-neural speech recognizers that can run compactly on edge devices. However, E2E ASR models are prone to overfitting and have difficulties in generalizing to unseen testing data.
Various techniques have been proposed to regularize the training of ASR models, including layer normalization, dropout, spectrum data augmentation and speed distortions in the inputs.
In this work, we present a simple yet effective noisy training strategy to further improve
the E2E ASR model training.
By introducing random noise to the parameter space during training, our method can produce smoother models at convergence that generalize better.
We apply noisy training to improve both dense and sparse state-of-the-art Emformer models and observe consistent WER reduction. 
Specifically,  when training Emformers with 90\% sparsity, 
we achieve 12\% and 14\% WER improvements on the LibriSpeech Test-other and Test-clean data set, respectively. 
\end{abstract}

%% file: tex/intro.tex
\section{Introduction}
\label{sec:intro}

Automatic Speech Recognition (ASR) enables fast and accurate transcriptions of voice commands and dictations on edge devices;
examples of on-device ASR applications include dictation for Google keyboard~\cite{schalkwyk_2019,he2019streaming}, voice commands for Apple Siri~\cite{vincent_2021}, and Amazon Alexa~\cite{liu2021exploiting}, etc. Past work developed streaming End-to-End (E2E) all-neural ASR models that run compactly on edge devices~\cite{he2019streaming, kim2019attention, shangguan2019optimizing,kim2020review}. These neural-network based E2E ASR models, however, are prone to overfit the training data,
and suffer from large performance degradation on noisy or unseen testing data, e.g., \cite{narayanan2019recognizing} 

Researchers have proposed many techniques to regularize the training of neural network-based models, which has shown to significantly improve the generalization of ASR models.  
Notable examples include layer normalization~\cite{Ba2016Layer}, early stop and dropout~\cite{zhou2017improved, gal2016theoretically}, audio reverberation and background noise simulations~\cite{kim2017generation}, spectrum data augmentation~\cite{Park2019,Park2020SpecAug}  
and speed perturbation of the audio inputs~\cite{ko2015audio,cui2015data}, etc.

In this work, we analyze a simple yet effective noisy training strategy for E2E streaming ASR model training via regularization. By introducing random noise to the parameter space during training, we force the ASR models to be smoother at convergence to achieve better generalization~\cite{an1996effects}. We select a strongly regularized state-of-the-art E2E ASR model -- an Enformer-based recurrent neural network transducer (RNN-T) model~\cite{shi2021emformer} and demonstrate how the proposed method can help to reduce the model word error rate (WER) on noisy testing dataset.

As E2E streaming ASR models are often constrained by the hardware resources when deployed on-device, pruning is widely adopted to simultaneously reduce the model computation and parameters. Hence, besides applying the proposed noisy training on the dense RNN-T model (73 MB), we also evaluate its effectiveness on several sparse RNN-T models, whose sizes range from 45 MB to 20 MB. We observe consistent WER reduction across all model sizes. In fact, we find noisy training more effective in improving downsized or sparse RNN-T models compared to the baseline models. The overall contribution of the paper can be summarized as follows:

\begin{itemize}
    \item For the first time, noisy training is applied to Enformer-based RNN-T models and systematically studied within an on-device E2E RNN-T framework.
    \item Noisy training is studied for both dense and sparse models across a wide range of model sizes. While consistent improvement is observed, we demonstrate 12\% and 14\% WER reduction on the LibriSpeech Test-other and Test-clean dataset for the 90\% sparse model. The experiments and ablation study help shed insights into the regularization of training small, on-device ASR models.
\end{itemize}

%% file: tex/method.tex
\section{Streaming E2E ASR}\label{sec:streamingASR}

In this work, we focus on streaming E2E ASR models, also known as as incremental recognizers -- as users speak, partial results already start to surface in real-time before the ASR finalizes its transcriptions. Streaming ASR models appear faster and more natural to users~\cite{aist2007incremental} and thus are popular for deployment on edge devices. They potentially reduce the latency of downstream models in a pipeline, such as in a real-time translation pipeline or a natural language understanding pipeline, because partial results can be fed downstream before the ASR finalizes its hypotheses~\cite{ShangguanKHMB20}.

Given a collection of data $\D = \{(x, y)\}$ with $x$ the acoustic signals and $y$ the corresponding text labels.
ASR models are often trained by maximizing the log-likelihood of the alignment sequence,
\begin{align}
    \L( w; \D) := \E_{(x, y)\sim \D} \bigg[ \log~p(y | x,  w))\bigg], 
    \label{eq:log_prob}
\end{align}
where 
$p(y| x, w)$ denotes the predictions of a ASR model and 
$w$ is the model parameters.
We use the RNN-transducer loss to compute $p$~
\cite{graves2012sequence}.
In this work,
we also adopt alignment-restriction RNN-T loss to speed up model training on GPUs~\cite{mahadeokar2021alignment}. 

This optimization in equation~\eqref{eq:log_prob} is prone to over-fitting. Empirically, we observe that the WER of training data is much smaller than that of the testing set even in the presence of other strong regularization techniques. Similar behavior is also reported by~\cite{2020GhodsiStateless} in the context of low-resource training data. We observe that the RNN-T losses and WER differences between training set and testing set are more obvious with compressed or sparsity-pruned E2E ASR models than large, dense models.

\begin{figure*}[ht]
    \centering
    \begin{tabular}{cc}
     \includegraphics[height=0.15\textwidth]{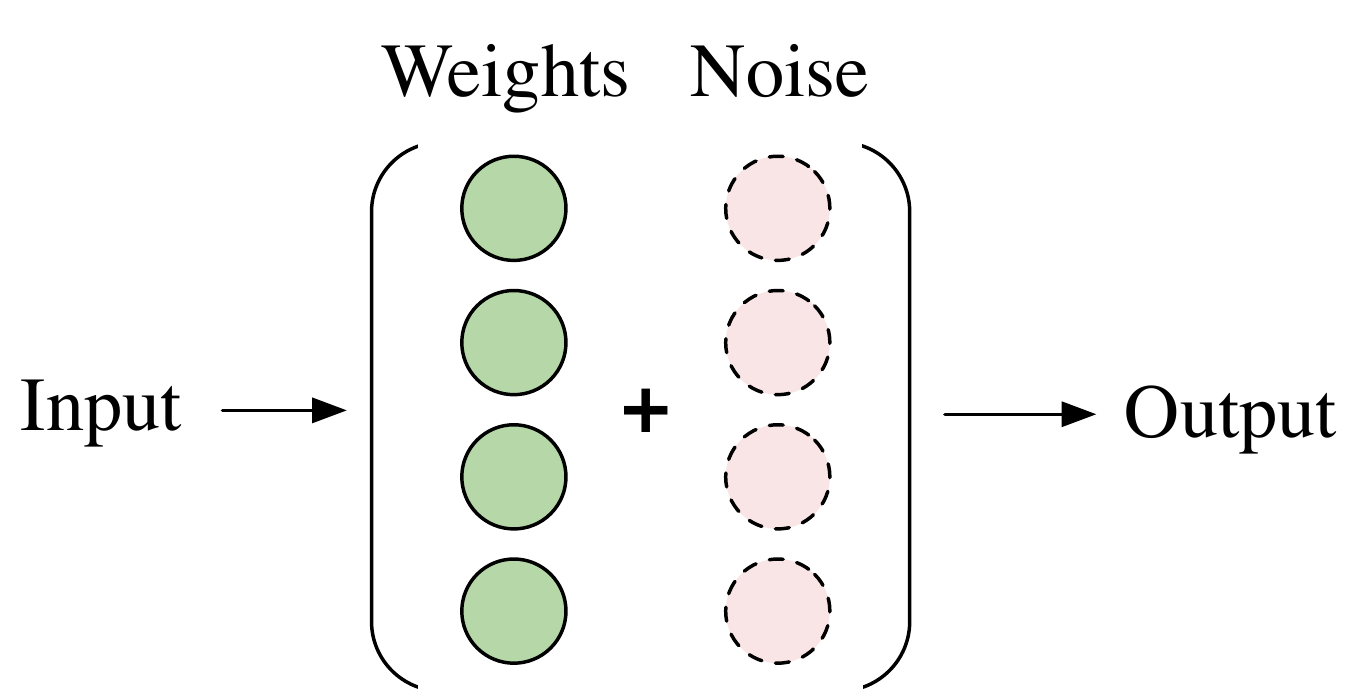} & 
      \includegraphics[height=0.15\textwidth]{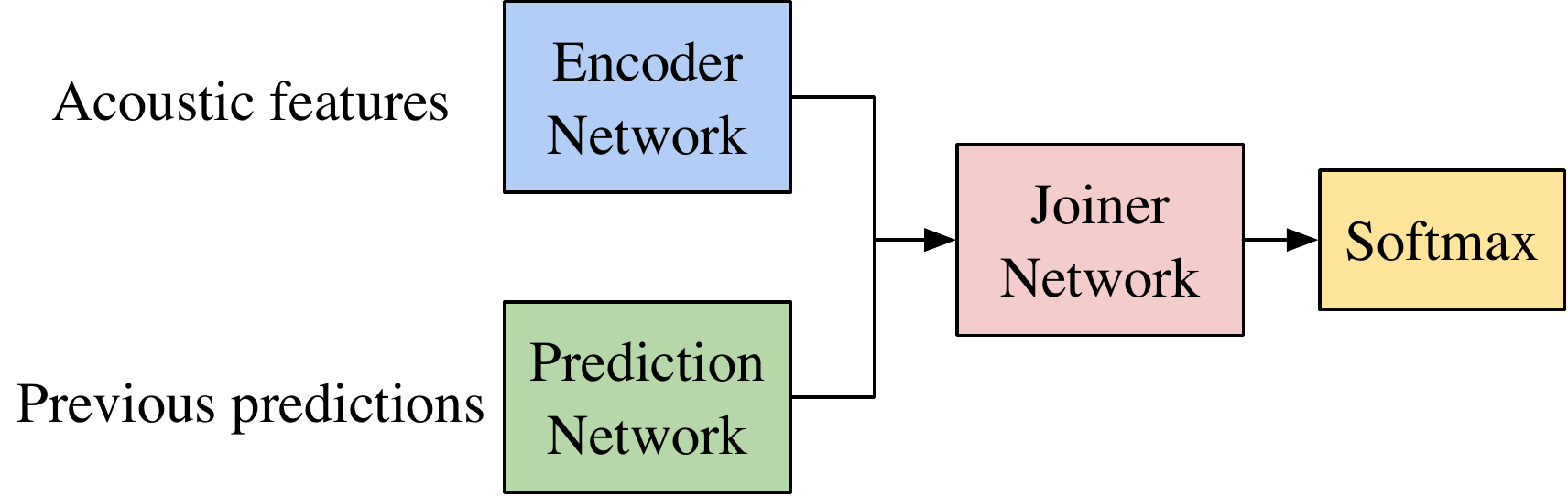} \\
     (a) parameter noise injection &  (b) model overview  \\
    \end{tabular}
    \caption{Overview of our noisy training method, and an illustration of the RNN-T model framework in this work.}
    \label{fig:overview}
\end{figure*}

\subsection{Noisy training}\label{sec:largemargin}
We investigate a simple method to effectively alleviates 
over-fitting when training ASR models.
Our method works by introduce random noise to the parameter space during training,
\begin{align}
     & \min_w \E_{\epsilon \sim p(\epsilon)} \L(w; \epsilon, \D) + \frac{\lambda}{2}\pp w \pp_2^2, \label{eq:main_noisy}~~~\text{where} \\
     &\L(w; \epsilon, \D) = \E_{(x, y)\sim \D} \bigg[ \log~p(y|x, w+\epsilon) \bigg]  \nonumber \\
     & p(\epsilon) = \N(0, \alpha^2). \nonumber
\end{align}
Here $p(\epsilon)$ is a noise distribution.
As shown theoretically in ~\cite{an1996effects}, 
this noisy training strategy effectively avoids 
sharp local minima at convergence.
Compared with standard MLE training in equation~\eqref{eq:log_prob},
optimizing a noise-perturbed objective yields smoother neural networks as solutions that generalize better. 
We refer the reader to  ~\cite{an1996effects} for more theoretical treatments. 

We set $p(\epsilon)$ as zero centered Gaussian distributions with variance $\alpha^2$ for simplicity. 
The squared term on $w$ is introduced to prevent $w$ from growing arbitrary large and hence, cancels out the regularization effect from the noise.
We set $\lambda =0.1$ as default throughout the paper.

Note that the exact computation of equation~\eqref{eq:main_noisy} is intractable.
In practice, we approximate equation~\eqref{eq:main_noisy} 
with a single Monte Carlo sample $\epsilon_0$ from $p(\epsilon)$
and  a random mini-batch $B$ from $\D$.
More specifically, at each training step,
we compute the gradient of $w$ as follows:
\begin{align}
\nabla_w \L(w; \epsilon_0, B), ~~~\text{with}~\epsilon_0 \sim \N(0, \alpha^2)
\label{eq:gradient}
\end{align}
We summarize our noisy training pipeline in Algorithm~\ref{alg:main}. See Figure~\ref{fig:overview} (a) for an illustration.

\begin{algorithm}[t]
\caption{Training ASR models with parameter space noise injection}
\label{alg:main}
\begin{algorithmic}[1]
\STATE {\bf Input}: Given a ASR model $f(w)$ with $w$ as its parameters; noise distribution $p(\epsilon)$
\WHILE{not converge}
    \STATE Sample a mini-batch of training data $B$
    \STATE Sample noise $\epsilon$ from the noise distribution $p(\epsilon)$
    \STATE Compute gradients $g = \nabla_w\L(w+\epsilon; B)$ according to equation~(\ref{eq:gradient}); and update $w$ using gradients $g$. 
\ENDWHILE
\end{algorithmic}
\end{algorithm}

\subsection{Connections to Variational inference}
Past works point out that Bayesian inference with neural networks improves model generalization, accuracy and calibration~\cite{DBLP:journals/corr/abs-2002-08791, DBLP:journals/corr/abs-2007-06823}. 
For ASR, exact Bayesian inference requires sampling from a Bayesian NN posterior, which is computationally intractable. 
Variational inference provides a computationally efficient tool that 
enables fast posterior approximation through optimization~\cite{David2017VI}.
Similar to Graves' formulation in~\cite{graves2011practical}, our noisy training could be viewed as finding a Gaussian proposal distribution over real value weight parameters $w$, $q_\theta(w)$, to approximate the posterior distribution of weights over the given dataset $\D$, $p(w\mid \D)$.

In variational inference, finding the best approximation of $p(w\mid \D)$ could then be formulated as minimizing the Kullback-Leibler (KL) divergence between $q_\theta(w)$ and $p(w\mid \D)$ as follows:
\begin{align}
\min_q \KL(q_\theta(w)~\pp~p(w \mid \D)). \label{eq:kl}
\end{align}
Furthermore, minimizing equation~\eqref{eq:kl} is equivalent to maximizing the following evidence lower bound (ELBO)~\cite{blei2017variational}:
\begin{align}
& \max_\theta\bigg\{ \mathrm{ELBO}(\theta) := -\KL(q_\theta(w) \pp p(w)) - \L_{\D}(\theta) \bigg\},~~~ \text{with} \label{equ:elbo}\\
& \L_{\D} = \E_{(x, y)\sim \D}~\E_{q_\theta(w)} \bigg[ \log p(y\mid x, w)\bigg],  \nonumber
\end{align}
where $\L_{\D}(\theta)$  is the expected log-likelihood.
Assume a simple Gaussian prior, e.g., $p(w) = \N(0, 1/\lambda)$~($\lambda>0$),
and assume $q_\theta(w)$ a fully factorized Gaussian, e.g., 
$q_\theta(w) = \N(\theta, \alpha^2)$ with $\alpha ^2$ as a constant.
In this way, the $\KL$ term can be integrated analytically, 
\begin{align}
\KL(q_\theta(w) \pp p(w)) \propto -\frac{\lambda}{2}\pp \theta \pp^2.
\label{eq:reg}
\end{align}
Substituting equation~\eqref{eq:reg} into equation~\eqref{equ:elbo}, we have
\begin{align}
\max_\theta \bigg\{ \mathrm{ELBO}(\theta) := -\frac{\lambda}{2}\pp\theta\pp^2 - \L_\D(\theta) \bigg\}. \label{eq:final_elbo}
\end{align}
It is easy to see that 
maximizing over $\theta$ as in equation~\eqref{eq:final_elbo} gives the same optimization objective as minimizing $w$ as in equation ~\eqref{eq:main_noisy}.

\subsection{Implementation Details}
Note that our method requires specifying a noise distribution $p(\epsilon)=\N(0, \alpha^2)$ (see equation~\eqref{eq:main_noisy}) to control the exploration in the parameter space, 
Intuitively, 
a large $\alpha$ might yield an optimization objective that is hard to converge;
On the other hand, a small noise might cannot sufficient regularize the training, and therefore yielding marginal benefits.
In this work, we find it is most effective to set $\alpha$ adaptively
according the magnitude of the weights during training. 

Specially, 
consider a linear layer with weights
$w\in \R^{d_{in}\times d_{out}}$, 
where $d_{in}$ represents the input dimension and
$d_{out}$ denotes the output dimension.
For each column of the weight matrix $w_j \in \R^{d_{in}}$, 
we perturb $w_j$ as follows, 
\begin{align}
    w_j  = w_j + \tilde{\alpha}\epsilon_j,~~\epsilon_j \sim \N(0, 1), \tilde{\alpha} = 
\alpha\frac{\pp w_i \pp_2}{\pp \epsilon_j \pp_2}, \epsilon_j \in \R^{d_{in}}, \label{eq:noise_scale}
\end{align}
During training, we treat $\tilde{\alpha}$ as constants and stop the gradients back-propagating through $\tilde{\alpha}$.

%% file: tex/exp.tex
\section{Experiments}\label{sec:results}
Our experiments are designed and reported in three parts:
in Section~\ref{subsec:denseEmformerResult} we apply our method to improve the training of RNN-T models, with different number depths and widths in the RNN-T Emformer-based encoder network; in Section~\ref{subsec:sparseEmformerResult}, we show the effectiveness of noisy training on sparsity-pruned ASR models; and finally in 
Section~\ref{subsec:ablationEmformerResults}, we provide extensive ablation studies to study the impact of the noise scale and the location to inject noises on the model's performance.

\paragraph{Dataset and data augmentation} 
We use the LibriSpeech 960h corpus for experiments~\cite{panayotov2015librispeech}. 
To enforce strong regularization from data augmentation, 
we perturb the input audio speed with ratio 0.9, 1.0 and 1.1 using techniques in~\cite{ko2015audio}.
We then extra 80-dimensional logMel features using a sliding window of 25ms and step of 10ms over the input audio. To regularize the model implicitly, we further apply spectrum data augmentation~\cite{Park2019} with Frequency mask parameter (F=27) and (T=10) Time masks with maximum time-mask ratio (p = 0.05).

\paragraph{Model architecture}
We use the recurrent neural network transducer (RNN-T) framework to represent E2E ASR models in this work. A RNN-T model typically contains three components: a encoder network, a prediction network and a joiner network. The encoder network converts frame-wise acoustic input into a high level vector representation; the prediction network acts as a language model that converts previously predicted non-blank tokens into a high level representation; the joiner combines encoder and prediction network output and applies a softmax to predict the next token, including a \textit{blank} token. We use a simplified Emformer cells~\cite{shi2021emformer} to build the encoder model of our RNN-T models. We use Long Short Term Memory (LSTMs) cells to build the predictor. We provide a more detailed description of our model architecture in Section~\ref{sec:emformer}.
We refer readers to~\cite{graves2012sequence,he2019streaming} for a more explanation on RNN-T. 
Additionally, we show the RNN-T structure used in this work in Figure~\ref{fig:overview}(b). 

We explicitly regularize all RNN-T models in this work by add extra modules, which lead to auxiliary losses to the intermediate RNN-T encoder layers as suggested in~\cite{liu2021improving}.

\paragraph{Regularization settings} 
We leverage state-of-the-art regularization
techniques to build a strong baseline training pipeline.
To summarize, we implement popular regularization techniques, including speech perturbation, spectrum data augmentation, layer normalization for all weight matrices in the RNN-T, and residual connection within the Emformer cells (see Section~\ref{sec:emformer}). We apply dropouts~\cite{srivastava2014dropout} of ratio 0.1 to the Emformer weight matrices, and dropouts of ratio 0.3 to all other linear and LSTM layers to reduce overfitting.

On top of the above-mentioned regularization techniques, we apply to noisy training to further boost the performance. We report the WERs on the LibriSpeech test-clean and test-other datasets. All WER results in this work is scored with \textit{NIST Sclite tool} without GLM grammar replacement~\cite{NISTtool}. 

\paragraph{Noisy training settings}
We add adaptive noise to the training following equation~\eqref{eq:noise_scale}. 
We set the noise scale $\alpha$ to be $0.01$ throughout the paper unless otherwise specified. We found this single setting performs well across all experimental setups.

\subsection{Improving Emformers}\label{subsec:denseEmformerResult}
In this part, we apply our noisy training to improve a number of Emformers with various model sizes. 

\paragraph{Settings}
Specifically, 
in Table~\ref{tab:emformer_shrink}, we denote Emformer-\textit{n}L as \textit{n} layers of Emformer used in the RNN-T transcription network. We also denote Emformer-20L(0.5x) as removing 1/2 of the hidden units in each of the 20 layers of Emformer cells in the RNN-T encoder network; in this way, the total number of parameters reduced is 75\% for the encoder.
All models are trained for 120 epochs with a batch size of 1024.

\paragraph{Results}
We find that our method achieves better WER compared to the baseline models without noisy training. The smaller the Emformer model, the more effective noisy training is in improving model performances. 

\begin{table}[ht]
    \centering
    \setlength{\tabcolsep}{1pt}
    \begin{tabular}{l|cll}
    \hline
   Method & \#Prams (M)  & Test-other & Test-clean  \\
   \hline
   Emformer-7L &  \multirow{2}{*}{35.7} & 13.0 & 5.2 \\
   + Noisy training  & & \bf 12.0 (-8\%) & \bf 4.6 (-12\%)\\
   \hline 
    Emformer-10L  &  \multirow{2}{*}{45.4}   & 11.5 & 4.5 \\ 
     + Noisy training & &  \bf 10.8 {\small (-6\%)} & \bf4.0 {\small (-11\%)} \\  \hline
     Emforer-14L &  \multirow{2}{*}{57.8} & 10.2 & 4.0 \\
   + Noisy training  & &  \bf 9.7 (-5\%) & \bf 3.8 (-5\%) \\ \hline
    Emforer-20L &  \multirow{2}{*}{76.7} & 9.9 & 3.8 \\
   + Noisy training  & &  \bf 9.5 (-4\%) & \bf 3.5 (-8\%) \\
     \hline \hline
      Emformer-20L (0.5x)  &  \multirow{2}{*}{29.1}  & 11.7 & 4.7  \\ 
      + Noisy training &  & \bf10.8 {\small (-8\%)} & \bf 4.3 {\small (-8\%)} \\
    \hline
    \end{tabular}
    \caption{Comparing baseline Emformer-based RNN-Ts training results with noisy training results. 
    }
    \label{tab:emformer_shrink}
\end{table}

\subsection{Improving Pruning aware training}\label{subsec:sparseEmformerResult}
Sparsity pruning introduces block-patterns of zeros inside the weight matrices of the neural networks. It allows the E2E ASR models to run compactly with low latencies on sparsity-friendly hardware. Sparse models have also been shown to outperform similar-sized compressed models without sparsity on speech and language modeling tasks~\cite{shangguan2019optimizing, Zhu2018Prune, pang2018compression}. In this section, we apply our noisy training method to the training of sparse Emformer-based ASR models. We use weight magnitude based pruning, and sparsify only the Emformer-based encoder network in an RNN-T, which occupies $>80\%$ of the size of the entire RNN-T model. 

\paragraph{Settings}
Each model in Table~\ref{tab:emformer_sparse} is trained from scratch with pruning-aware training. We use a sparsity block pattern of 8x1, and a cubic pruning schedule described in~\cite{Zhu2018Prune}:
\begin{align}
    s_t=s_f \bigg(1- \bigg(1-\frac{t-t_0}{n\Delta t} \bigg)^3\bigg),
    \label{equ:prune_interval}
\end{align}
where $t\in \{t_0, t_0+\Delta t, ..., t_o + n\Delta t\}$.
Here $s_t$ denotes the pruning ratio at training step $t$
and $s_f$ is the target sparsity ratio.
We starts pruning at step $t_0$ and gradually increase the pruning ratio during training. Meanwhile, $\Delta t$ represents the pruning frequency and $n$ denotes the number of pruning steps.
We set both $n$ and $\Delta t$ to be 256 for all sparse models.

Meanwhile, as it is more difficult to optimize sparsified models.
For each pruning-aware training setting with pruning ratio $v$, 
we linearly scale down its corresponding dropout ratio on the Emformer encoder from $0.1$ to $0.1 * (1-v)$. 

\paragraph{Results}
As shown in Table~\ref{tab:emformer_sparse}, 
our noisy training leads to consistent WER reduction on all the settings evaluated. 
In particular, 
the improvements become increasingly significant as 
we gradually increase the pruning sparsity. 
Specifically, for Emformers with 90\% sparsity, 
our method achieves 12\% and 14\%
WER reduction on the test-other and the test-clean data set, respectively, 
compared its correpoding basleine model;
for Emformer with 50\% sparsity, our results are comparable  
with the results from the standard baseline without noisy training. 

Additionally, compared with Emformer-10L in Table~\ref{tab:emformer_shrink},
our Emformer with 50\% sparsity achieves better WER while maintaining a similar model size.
Our results confirm the effectiveness of network pruning for ASR model compression.

\begin{table}[ht]
    \centering
    \setlength{\tabcolsep}{3pt}
    \begin{tabular}{l|cll}
    \hline
   Method & \#Prams (M) & Test-other & Test-clean   \\
   \hline 
    Sparse 50\%  & \multirow{2}{*}{45.1} &  10.7 & 4.0  \\ 
     + Noisy training & & \bf 10.1 {\small (-6\%)} & \bf 3.7 {\small (-8\%)} \\ 
     \hline
      Sparse 70\% & \multirow{2}{*}{32.4} &11.2 & 4.5  \\
      + Noisy training & & \bf 10.4 {\small (-7\%)} & \bf4.1 {\small (-9\%)} \\
    \hline
    Sparse 90\% & \multirow{2}{*}{19.6}  & 14.7 & 5.9 \\
      + Noisy training & &\bf 12.9 {\small  (-12\%)} & \bf 5.1 {\small  (-14\%)} \\
    \hline
    \end{tabular}
    \caption{Results for sparse Emformer models.
    }
    \label{tab:emformer_sparse}
\end{table}

\begin{figure*}[ht]
    \centering
    \setlength{\tabcolsep}{1pt}
    \begin{tabular}{ccc}
    \raisebox{3em}{\rotatebox{90}{Training loss}}
    \includegraphics[width=0.3\textwidth]{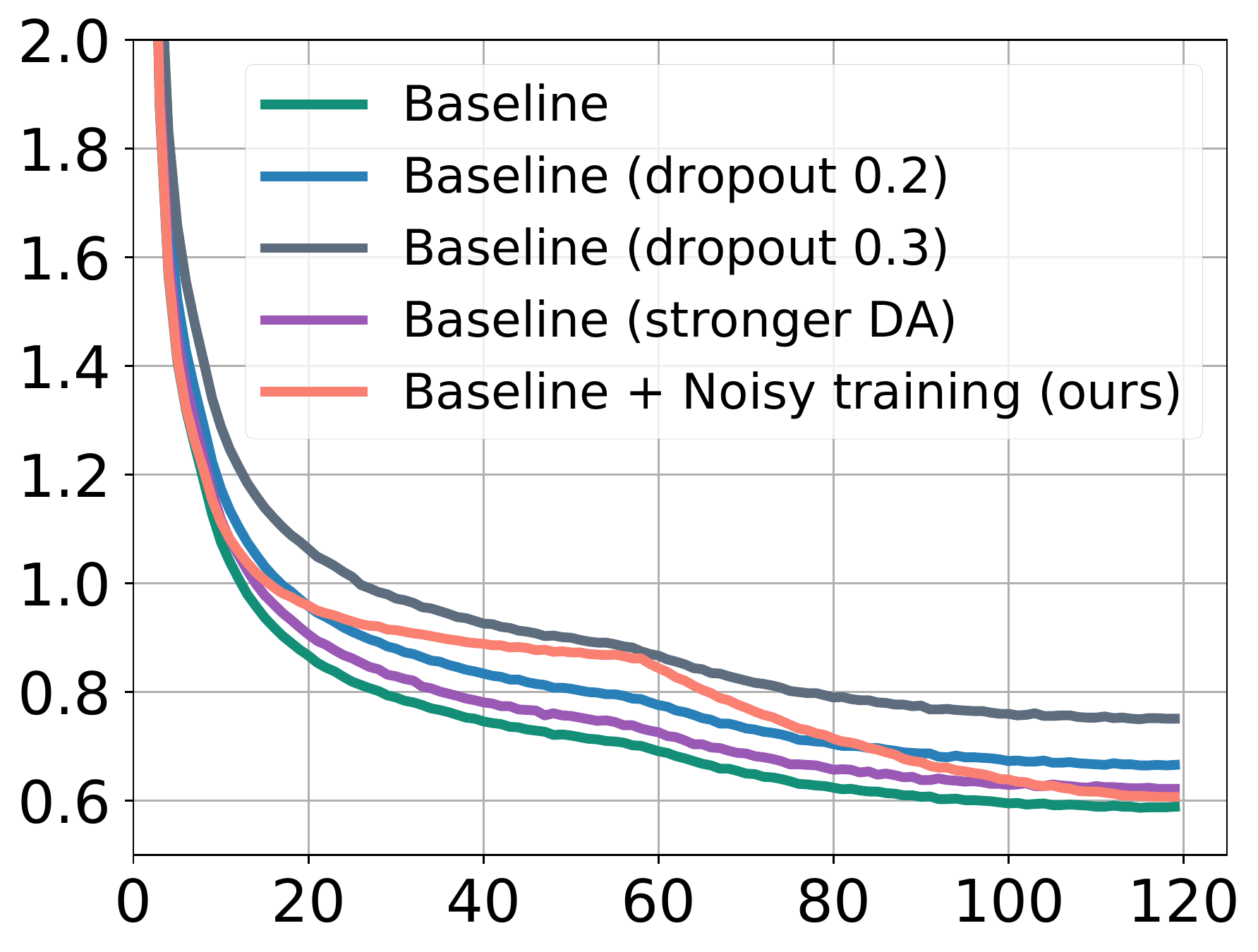} &   \raisebox{3em}{\rotatebox{90}{Validation loss}}
    \includegraphics[width=0.3\textwidth]{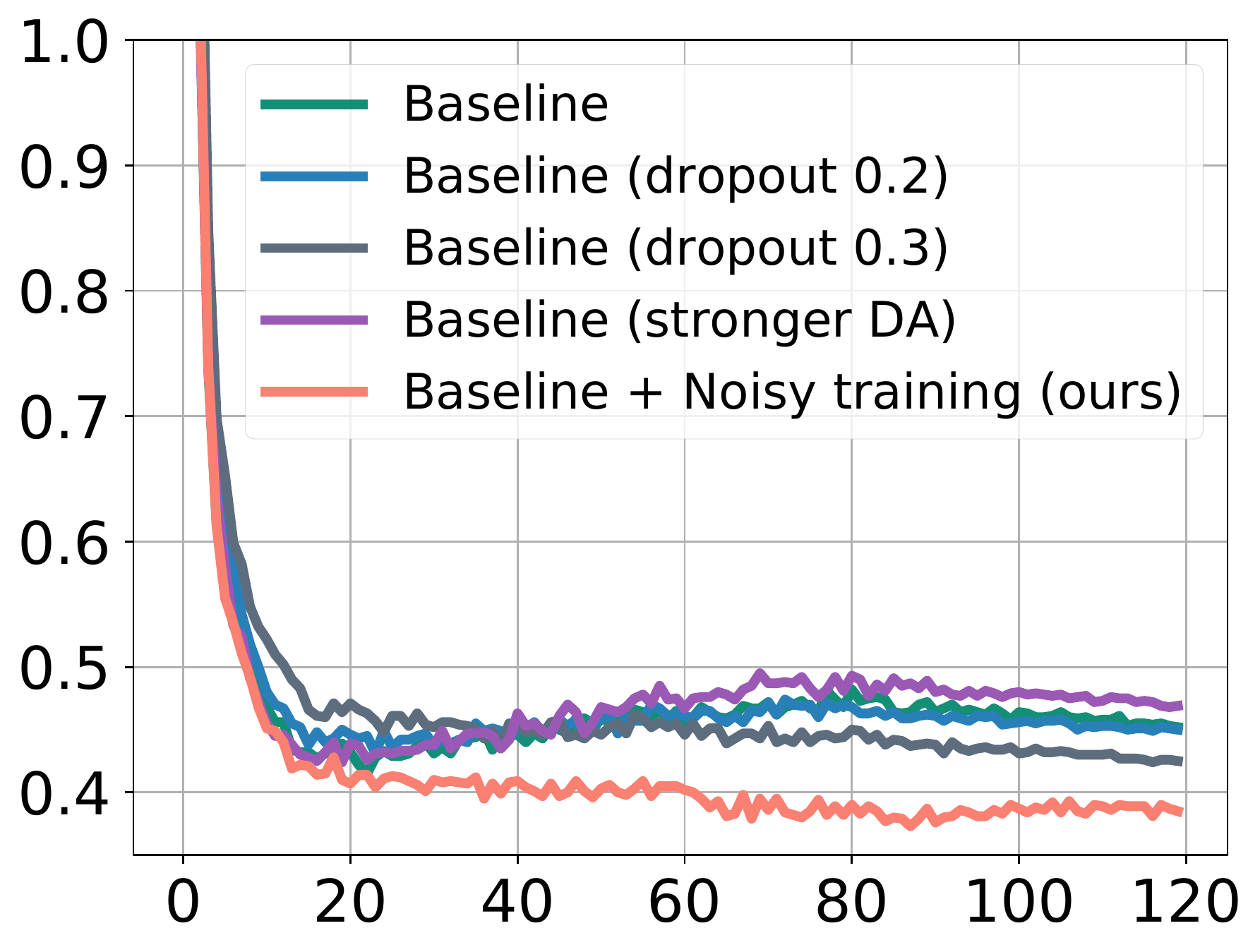} &
    \raisebox{3em}{\rotatebox{90}{Log Eigenvalues}}
    \includegraphics[width=0.3\textwidth]{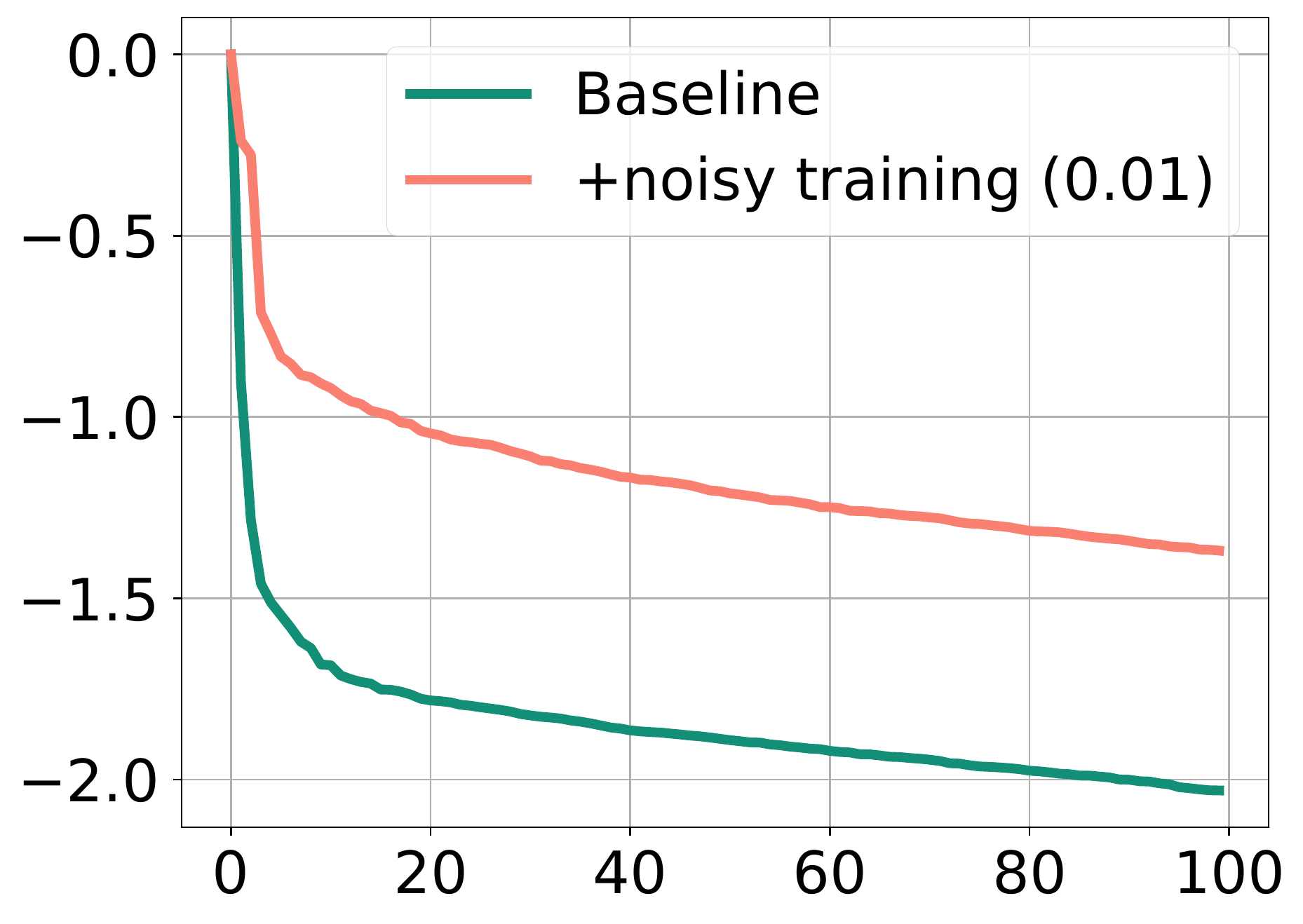}
    \\
    Training epoch & Training epoch & Index of Singular Values \\
    (a)  & (b) & (c) \\
    \end{tabular}
    \caption{(a) Training loss during training.
    (b) Validation loss during training. 
    Our method effectively alleviates over-fitting. 
    (b) Logarithmic scale singular
values of the embedding matrix. Note we normalize the singular values of each matrix so that the largest one is 1.}
    \label{fig:val_loss}
\end{figure*}

\subsection{Ablation Studies}\label{subsec:ablationEmformerResults}
We provide empirical results to show how the noise can be tuned, where noise should be added, and how overfitting can be further reduced in the baseline Emformer-20L model which already has larger dropout ratios and strong data augmentation.

\paragraph{Magnitude of noise} 
By sweeping the hyper-parameter of noise scales, we show in Table~\ref{tab:dense_20_noise_scale} that noisy training reduces model overfitting and results in better WER performances while the results are relatively insensitive to the magnitude of noise.

\paragraph{Location of noise}
By adding noise to each component of the RNN-T model separately, we see in Table~\ref{tab:noise_location} that noisy training is most effective when added to all parts of the RNN-T models; see an illustration of our model architecture in Figure~\ref{fig:overview} (b).

\begin{table}[ht]
    \centering
    \begin{tabular}{l|cc}
    \hline
   Method & Test-other & Test-clean \\
   \hline 
    Baseline &  9.9 & 3.8 \\ \hline
      + Noisy training (0.005) & 9.7 & 3.6 \\
      + Noisy training (0.01) & \bf  9.5  & \bf 3.5 \\
      + Noisy training (0.05) &  9.6 & 3.6 \\
      + Noisy training (0.1) & 9.8 & 3.6 \\
    \hline
    \end{tabular}
    \caption{
    Comparing results of noisy training with different noise magnitudes.
    The number in each parentheses denotes the noisy scale. 
    }
    \label{tab:dense_20_noise_scale}
\end{table}

\begin{table}[ht]
    \centering
    \begin{tabular}{l|cc}
    \hline
   Where to +noise & Test-other & Test-clean \\
   \hline 
   Baseline: none & 9.9 & 3.8 \\
    Baseline: all & \textbf{9.5} & \textbf{3.5} \\\hline
    Encoder  only  & 9.7 & 3.6 \\
    Predictor  only  & 9.8 & 3.6  \\
    Joiner  only &  9.7 & 3.6 \\
    \hline
    \end{tabular}
    \caption{
    WER results with respect to adding noise to different parts of the RNN-T model during training.
    }
    \label{tab:noise_location}
\end{table}

\paragraph{Logit space noise injection}
Inspired by past work on noise injection into the pre-softmax output logits~\cite{wang2019improving}, in addition to the parameter space noise injection, we further explore the idea of adding noise to the output logits of RNN-T model.

Table~\ref{tab:dense_20_logit_space_noise} shows that noise added to the logits does not help improve the model further. We hypothesize that both logit-space noise injection and weight noise injection achieves similar effects of regularization in the RNN-T model.

\begin{table}[t]
    \centering
    \begin{tabular}{lc|cc}
    \hline
   Model & Logit noise & Test-other & Test-clean \\
   \hline
   \multirow{2}{*}{Dense} & 0.01 & 9.5 & 3.6 \\
                          & 0.05 & 9.6 & 3.5 \\
    \hline
    \end{tabular}
    \caption{
    Adding Gaussian noise to the logit space cannot further improve the WER. ``Logit noise'' denotes the standard deviation of the Gaussian noise distributions used. 
    }
    \label{tab:dense_20_logit_space_noise}
\end{table}

\paragraph{Noisy Training compliments other regularization}
First, we explore whether increasing the strengths of dropouts or data augmentation could lead to the same improvements in the model's WER and reduce model overftting. In Table~\ref{tab:dense_20_large_reg}, we show results of models when we increase the dropouts on Emformer from 0.1 to 0.2 and 0.3. We also show results of the baseline model configuration, with stronger data augmentation by increasing the time-mask probability from 0.2 to 0.3. These changes lead to higher training loss, slightly lower validation loss, but higher WERs. 

\begin{table}[ht]
    \centering
    \begin{tabular}{l|cc}
    \hline
   Method & Test-other & Test-clean \\
   \hline 
   Dropout 0.2 & 9.9 & 3.9 \\
   Dropout 0.3 & 10.6 & 4.1 \\
   Stronger SpecAug & 9.9 & 3.8 \\ 
    \hline
    \end{tabular}
    \caption{
    Larger regularization to train the baseline.
    }
    \label{tab:dense_20_large_reg}
\end{table}

Noisy training compliments these existing regularization techniques, and is able to force the model to generalize better. We plot out the training loss and validation loss during training for the baseline model, models in Table~\ref{tab:dense_20_large_reg}, and the oisy trained model in figure~\ref{fig:val_loss}. Comparing with the baseline model, the noisy trained model converges slower at the beginning of the training but its training loss continues to decrease at a faster rate after 60 epochs. 
With noisy training, the validation loss is significantly lowered. 

To illustrate the regularization power of our noisy training, we follow work in~\cite{gao2019representation} and study the representation power of the learned Emformer-based RNN-T. 
Specifically, we visualize the weight matrix from the last linear layer of the joiner network, which has size $4097 \times 1024$. When trained with noisy training, the singular values distribute more uniformly, an indication that noisy trained embedding vectors fills a higher dimensional subspace.

%% file: tex/emformer.tex
\section{Additional details on Emformer-based RNN-T}
\label{sec:emformer}

\begin{figure}[t]
    \centering
    \begin{tabular}{c}
    \includegraphics[width=0.9\linewidth]{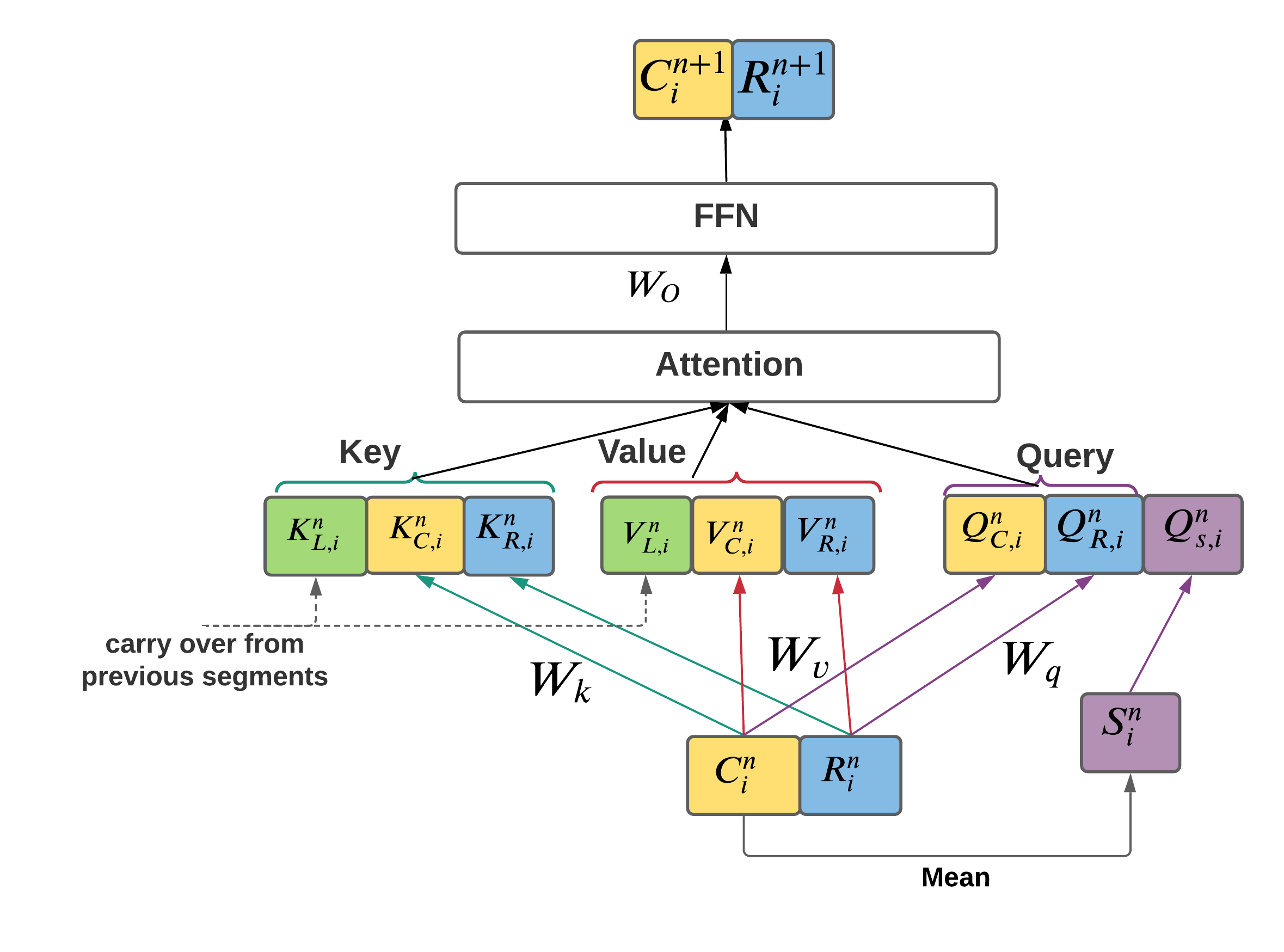}
    \end{tabular}
    \caption{Simplified Emformer Cell (no memory block) }\label{fig:EmformerCell}
\end{figure}

We build the RNN-T model encoder network with various layers of Emformer cells~\cite{shi2021emformer}, sandwitched between 2 linear projection layers. Emformer, an efficient variants of the transformer architecture~\cite{waswani2017attention}, is one of the state-of-the-art architecture for E2E streaming speech recognizers. Each Emformer cell has 8 attention heads, 512 hidden units and the attention feed-forward network of 2048 units. We do not use the ``memory bank" because in each step, the Emformer processes a long enough segment (160ms) of input. We thus simplify the Emformer with the the following formulation (using similar mathematical notation as in~\cite{shi2021emformer}): 
\begin{eqnarray}
[\hat{C}^n_i, \hat{R}^n_i] = LayerNorm([C^n_i, R^n_i])\\
K^n_i = [K_{L,i}^n, W_kC^n_i, W_kR^n_i]\\
V^n_i = [V_{L,i}^n, W_vC^n_i, W_vR^n_i]\\
Z_{C,i}^n = Attn(W_q\hat{C}^n_i, K_i^n, V_i^n) + C_i^n \\
Z_{R,i}^n = Attn(W_q\hat{R}^n_i, K_i^n, V_i^n) + R_i^n \\
Z_{Q,i}^n = Attn(W_q s_i^n, K^n_i; V^n_i)\\
\hat{X}^n_{i+1}=FFN(LayerNorm([Z^n_{C,i}, Z^n_{R,i}, Z^n_{Q,i}]))\\
X^{n+1}_i = LayerNorm(\hat{X}^{n+1}_i + [Z^n_{C,i}, Z^n_{R,i}, Z^n_{Q,i}])
\end{eqnarray}
where an input segment sequence is $C_i^n,...,C_{I-1}^n$ with i denoting the segment index and n denoting the layer index. $R_i^n$ refers to the right context segment; in this work we use only 1 right context segment. $K,Q,V$ refers to the key, query and value originally specified in transformer cells~\cite{waswani2017attention}. $W_k, W_q,W_v$ are the weight matrices associated with the computation for $K,Q$ and $V$ respectively. $Attn$ is the attention operation and $FFN$ refers to a feed-forward network. Figure~\ref{fig:EmformerCell} shows a visualization of these operations in the Emformer layer.

We build the predictor network of the RNN-T model with 3 layers of LSTMs with 512 hidden units, which are sandwiched between an linear Embedding layer and a linear output projection layer. Our LSTM layers are the same as shown in~\cite{shangguan2019optimizing}. A layer normalization is added to stabilize the hidden dynamics of the cell~\cite{Ba2016Layer}. The joiner network of the RNN-T contains one linear layer of size 1024, and a softmax that predicts the probabilistic distributions of 4097 tokens: 4096 pre-trained sentence pieces~\cite{kudo-richardson-2018-sentencepiece} and a \textit{blank} token.

%% file: tex/priorwork.tex
\section{Prior Works}

Gaussian weight noise injection into neural network training is not a new technique. Noisy training for recurrent neural networks has been discussed as early as in 1996 in~\cite{jim1996analysis}. The paper concluded that noisy training in feed-forward RNNs resulted in faster convergence and better generalization. 

Fast forward to 2011, Graves framed Gaussian weight noise injection during the training of neural networks in the context of variational inferences~\cite{graves2011practical}; Graves et al. in~\cite{graves2013speech} further applied Gaussian weight noise empirically into a speech model training process. There are two main differences between~\cite{graves2013speech} and this work. First, the noise was added once per training sequence to a Long Short Term Memory-based Acoustic model with phoneme targets and CTC loss. In our work, however, we apply noisy training in every forward step of the model, to a transformer-based E2E ASR model with RNN-T loss. Secondly, their best models in that work were first trained without noisy training to the best log-likelihood over the dev set, before being fine-tuned with noisy training further. In our work, we train the models from scratch with noisy training, simplifying the model training process.

Similarly, Shan et al. and Toshniwal et al. both applied Gaussian weight noise to the attention-based non-streaming Listen-attend-and-Spell (LAS) E2E model training~\cite{shan2018attention,toshniwal2018multilingual}. They mentioned noisy training but did not discuss in depth the extent of contribution of noisy training with respect to other regularizes, or the impact of noisy training on model of different sizes like we do.

%% file: tex/conclusion.tex
\section{Conclusion}
In this work, we analyze the impact of noisy training on a streaming, on-device E2E ASR model, trained with the RNN-T loss and existing strong regularization techniques. We systemically studies the ablation of noisy training with respect to the location and the magnitude of noises added. To support on-device compressed model deployment, this work specifically studies the impact of noisy training on transformer-based RNN-T models that are compressed or sparsity-pruned. We show that noisy training brings disproportionately more performance gain on smaller models, by reducing model overfitting, which could not be simply achieved via increasing the strengths of other regularizers such as drop out or data agumentation.